\pdfoutput=1

\documentclass[11pt]{article}
\usepackage[table,dvipsnames]{xcolor}
\usepackage[]{EMNLP2023}

\usepackage{times}
\usepackage{latexsym}

\usepackage[T1]{fontenc}

\usepackage[utf8]{inputenc}

\usepackage{microtype}

\usepackage{inconsolata}
\usepackage{amsmath}
\usepackage{color, colortbl}

\usepackage{csquotes}
\usepackage{graphicx}

\usepackage{array,booktabs}
\title{Adaptive Retrieval-Augmented Generation for Conversational Systems}

\author{
Xi Wang$^{1}$, Procheta Sen$^{2}$, Ruizhe Li$^{3}$, Emine Yilmaz$^{4}$ \\
$^{1}$University of Sheffield, $^{2}$University of Liverpool, \\ $^{3}$University of Aberdeen, $^{4}$University College London \\
\texttt{xi.wang@sheffield.ac.uk}$^{1}$, \texttt{procheta.sen@liverpool.ac.uk}$^{2}$, \\\texttt{ruizhe.li@abdn.ac.uk}$^{3}$, \texttt{emine.yilmaz@ucl.ac.uk}$^{4}$
}

\begin{document}
\maketitle
\begin{abstract}
Despite the success of integrating large language models into the development of conversational systems, many studies have shown the effectiveness of retrieving and augmenting external knowledge for informative responses. Hence, many existing studies commonly assume the \textit{always} need for Retrieval Augmented Generation (RAG) in a conversational system without explicit control. This raises a research question about such a necessity. In this study, we propose to investigate the need for each turn of system response to be augmented with external knowledge. In particular, by leveraging human judgements on the binary choice of adaptive augmentation, we develop \textit{RAGate}, a gating model, which models conversation context and relevant inputs to predict if a conversational system requires RAG for improved responses. We conduct extensive experiments on devising and applying \textit{RAGate} to conversational models and well-rounded analyses of different conversational scenarios. Our experimental results and analysis indicate the effective application of \textit{RAGate} in RAG-based conversational systems in identifying system responses for appropriate RAG with high-quality responses and a high generation confidence. This study also identifies the correlation between the generation's confidence level and the relevance of the augmented knowledge.


\end{abstract}

\section{Introduction}

\begin{figure}
    \centering
    \includegraphics[width=0.9\columnwidth]{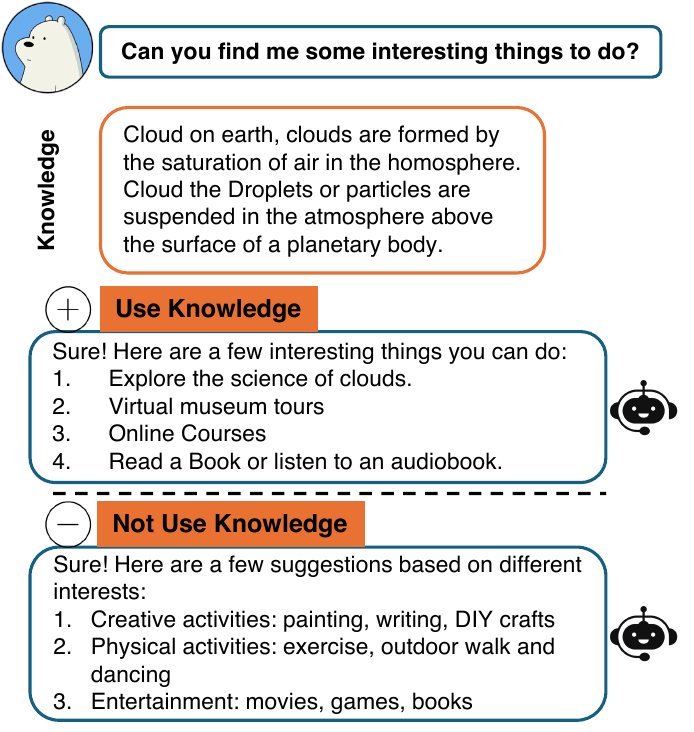}
    \caption{Example conversation when generating a response with or without a knowledge snippet using a language model (GPT-4 in this example). }
    \label{fig:illusrative_conv}
\end{figure}

Recently, the advancement of Large Language Models (LLMs) has significantly improved conversational systems, enabling the generation of natural and high-quality responses~\cite{ni2023recent}. Despite these advancements, recent studies have identified several limitations on the simple use of LLMs to address conversational tasks~\cite{onoe2022entity,huang2021plato,ren2018towards}. These limitations include the lack of up-to-date knowledge~\cite{onoe2022entity}, the generation of non-factual or hallucinated content~\cite{huang2021plato}, and restricted domain adaptability~\cite{ren2018towards}. These issues can hinder the development of conversational agents with satisfactory user experience. To address these identified challenges, a common approach is to retrieve and augment LLMs with external knowledge to enhance the conversational response, making them more accurate, reliable, and adaptable to different domains~\cite{zhao2020knowledge,lian2019learning,ye2024boosting}. For example,~\citet{shuster2021retrieval} demonstrated that using a dense retrieval model (DPR)~\cite{karpukhin2020dense} to retrieve relevant knowledge for augmentation can significantly reduce the hallucination rate, according to a corresponding human evaluation. Similarly,~\citet{yang2020graphdialog} showed that leveraging a graph-structured knowledge base can boost the reasoning ability and domain generalisability of task-oriented conversational agents. These achievements of knowledge-augmented techniques highlight a promising direction for enhancing conversational agents and address the current limitations. 

However, while implementing retrieval augmentation to a conversational system for improved response, we question the necessity of knowledge augmentation for every turn of system responses. To develop effective human-computer conversations, it is essential to provide factual and relevant responses, offer \textit{appropriate} amount of information, and not unnaturally drive and shift the conversation to non-relevant topics \cite{kasirzadeh2023conversation,miehling2024language}. We argue that overusing external knowledge could result in system responses against these core criteria. Figure~\ref{fig:illusrative_conv} presents a conversation example that shows how the system response to a generic user utterance about suggesting activities can vary with and without augmented knowledge. The knowledge-augmented system response is being information conditioned with limited diversity and assuming specific user preferences.
In contrast, without the addition of external knowledge, the system response is more diverse and natural in this early stage of a conversation. This indicates that misusing external knowledge can lead to problematic system responses and a negative user experience. 

To address this, we investigate an adaptive retrieval-augmented generation solution for effective conversational systems. In particular, motivated by the gate function in long-short term memory models \cite{graves2012long}, which explicitly controls the use of input and memory, we propose a binary knowledge gate mechanism, called \textit{RAGate}, to manipulate the use of external knowledge for a conversational system. To model the conversation context and accurately estimate the need for augmentation, we leverage the human labels as ground truth and develop RAGate by exploring the use of recent advanced language models or constructing attention neural gate models. To validate the effectiveness of RAGate, we conduct extensive experiments on an annotated Task-Oriented Dialogue (TOD) system dataset, KETOD, that builds upon the SGD dataset with TOD-spanning
16 domains, such as Restaurant and Weather. The experimental results show that RAGate enables conversational systems to efficiently use external knowledge at appropriate conversation turns, producing high-quality system responses. In particular, by modelling the uncertainty and confidence level of the system -- which correlates with the likelihood of hallucinated output \cite{varshney2023stitch} -- we show that the "always" augmentation of external knowledge can significantly increase generation uncertainty and the risk of hallucination. After applying RAGate, we can effectively control the conversation system to make confident and informative responses. In addition, by varying the use of knowledge snippets in different relevance levels, we also observe the positive correlation between the calculated confidence score and the relevance of augmented knowledge, which can be valuable for many future studies. 

\begin{figure*}
    \centering
    \includegraphics[width=1.2\columnwidth]{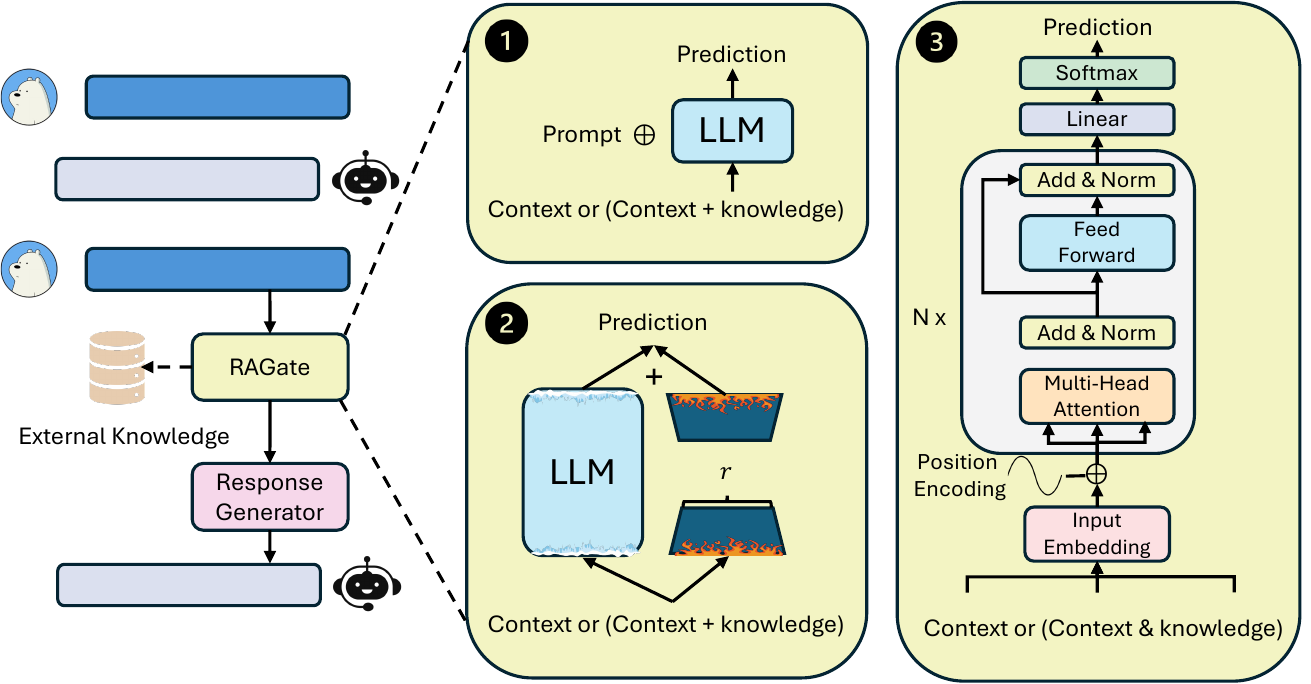}
    \caption{\textit{RAGate} variants for implementing the gating function. The three variants are the prediction with pre-trained language models after prompting (1), after parameter-efficient fine-tuning (2), and with a multi-head attention encoder (3).}
    \label{fig:ragate}
\end{figure*}

\section{Related Work}
In the pipeline of knowledge-augmented generation for a conversation system, two main components are identified: the knowledge retriever and the response generator. Existing studies have improved conversational responses to different extents by improving one or both components~\cite{li2022knowledge,komeili2022internet,wang2024unims}. 

\textbf{\textit{Knowledge Retrieval:}} Several studies have explored the use of dense passage retrieval techniques \cite{lewis2020retrieval,karpukhin2020dense} and public search service for effective retrievers \cite{li2022knowledge}. For example, \citet{li2022knowledge} retrieved Wikipedia passages through a database interface and then ranked them according to statistical relevance, calculated by TF-IDF, or semantic relevance as per cosine similarity. Similarly, \citeauthor{komeili2022internet} used a search engine API to retrieve relevant knowledge but first transformed the dialogue context into a natural search query using an encoder-decoder model before searching. 

\textbf{\textit{Joint Optimisation of Retriever and Generator:}} On the other hand, another thread of research studies has explored joint optimisation approaches. For instance,~\citet{shi2023dual} introduced a retriever-generator architecture that aims to improve the performance of Task-Oriented Dialogue (TOD) systems by using a dual-feedback mechanism. The retriever identifies relevant knowledge from a database, while the generator uses this information to create appropriate system responses. The feedback from the generator is further used as pseudo-labels to train the retriever to select pertinent information.~\citet{shen2023retrieval} introduced a training method based on maximal marginal likelihood. This method jointly optimise a perceptive retriever and the response generation in a feedback loop. The proposed approach incorporates meta-knowledge, which guides the generator to improve the utilisation of knowledge and, consequently, the quality of the generated responses. ~\citet{kang2023knowledge} further advance the retriever by proposing SUbgraph Retrieval-augmented GEneration (SURGE), which employed a graph neural network (GNN)-based context-relevant subgraph retriever. SURGE incorporates contrastive learning to optimise the latent representation space, ensuring that generated texts closely resemble the retrieved subgraphs. 

Despite the richness of existing retrieval-augmented generation techniques for conversational systems, they commonly hypothesise that every conversation turn needs external knowledge. However, the necessity of augmenting every turn of the conversation with external knowledge remains questionable. A relevant thread of work that aims to answer this question is the introduction of the knowledge-seeking turn detection task using the DSTC-9 dataset \cite{kim2020beyond}, and the follow-up studies, such as \cite{hong2023knowledge,jin2021towards}. However, this task is to identify the turns in conversations injected by human workers about knowledge enquiry instead of identifying the system responses that require knowledge augmentation for improvements. This research gap highlights the value and novelty of this study, which investigates the adaptive use of retrieval-augmented generation for advanced conversational systems.

\section{Methodology}\label{sec:methodology}

\subsection{Problem Formulation}
This study addresses the challenge of effectively identifying conversation turns that require augmentation of external knowledge. In particular, we aim to develop a gate mechanism that dynamically determines when to search for external knowledge to ensure natural, relevant and contextually appropriate responses. First, we define the task of user-system conversation. Let $D = \{d_1, d_2, ...,d_{|D|}\}$ be a set of user-system dialogues, and each dialogue $d$ comprises a sequence of interactions between users and systems (i.e., $d = \{u_0, s_0, u_1, s_1, ..., u_T, s_T\}$) with varying lengths. Here, $u_t$ and $s_t$ denote the user utterance and system response at the $t$-th turn, respectively. The conversational context up to turn $t$ can be formulated by aggregating the previous user-system interactions, i.e., $c_t = {u_0, s_0, .., u_t}$. With this context information $c_t$, the conversation system can augment it with a list of retrieved external knowledge, $e_{t,k}$, where $k$ represents the ranking cutoff for the retrieved knowledge. Hence, the binary gate mechanism proposed in this study, deciding the knowledge augmentation, can be formulated as $f(c_t) = \{0,1\}$ or $f(c_t, e_{t,k}) = \{0,1\}$ if the external knowledge $e_{t,k}$ is considered. Then, the follow-up response generation function $g(\cdot)$ can be formulated as follows:
\begin{equation}
    g(\cdot) = \begin{cases}
        g(c_t, e_{t,k}) & \text{if $f(c_t)$ or $f(c_t, e_{t,k})$} \\
        g(c_t) & \text{otherwise}.
    \end{cases}
\end{equation}
Hence, by evaluating and estimating the necessity of augmenting with external knowledge, we dynamically update the conversational response generation accordingly.

\subsection{RAGate Gate Mechanism}
To effectively estimate the need to use external knowledge and implement adaptive retrieval augmented generation for a conversation system, we introduce our proposed gate mechanism, RAGate, that uses the conversational context and, optionally, the retrieved external knowledge to predict the binary choice of using external knowledge. In particular, we explore three RAGate variants that are implemented by the use of Large Language Models (LLMs) with devised prompts, with parameter efficient fine-tuning (e.g., QLoRA \cite{dettmers2024qlora}) and the construction of an end-to-end multi-head attention encoder. This exploration is motivated by the recent advancement of transformer-structured neural models in natural language processing. In Figure \ref{fig:ragate}, we illustrate the application of RAGate and its three variants. We describe each of these three variants to clarify the use of RAGate:

\textit{\textbf{RAGate-Prompt: }} As denoted by \citet{arora2022ask}, a language model can effectively adapt to new tasks by using a natural language prompt that explains the process to address the tasks without extra training. Hence, we can formulate 
a gate function $f(\cdot)$ as $f(y|c_t) = f(y|\Theta, c_t, p)$, where $\Theta$ denotes the used language model with its pre-trained weights and $p$ is the devised natural language prompt. Alternatively, if the retrieved knowledge is also involved in prediction, we have $f(y|c_t) = f(y|\Theta, c_t, e_{t,k}, p)$. Specifically, we explore two types of prompts: zero-shot and in-context learning. Zero-shot prompts describe the task that uses the conversational context and, optionally, the retrieved knowledge to generate a response with binary feedback. As for the in-context learning prompts, we augment the zero-shot prompts with illustrative examples. We show the set of prompts in Appendix \ref{sec:ragate-prompt}. 

\textit{\textbf{RAGate-PEFT: }} Despite the high adaptability of the language model with devised prompts, we further explored the use of instruction tuning on language models with a parameter-efficient fine-tuning method (i.e., QLoRA \cite{dettmers2024qlora}) to meet the goal of an effective gate function. QLoRA is built upon the known Low-rank Adapter (LoRA) \cite{hu2021lora}, which keeps the pre-trained weight matrix $W_0$ frozen and addresses the gradient updates of the weight matrix $\Delta W$ through low-rank approximation (i.e., $\Delta W = BA$, where $B$ and $A$ are the result of lower-rank decomposition on $\Delta W$). Hence, the forward pass during the model training can be updated from $h = W_0x+\Delta Wx$ to $h = W_0x + BAx$. QLoRA \cite{dettmers2024qlora}, which is used in this study, further quantises the language model into a 4-bit NormalFloat data type and leverages the page-to-page transfer between the CPU and GPU to further avoid memory spikes. To implement RAGate-PEFT, we format the train data with devised instructions, joined with paired inputs and outputs for developing parameter-efficient fine-tuned large language models. In particular, we provide a set of instruction-input-output triples for model training. The input can vary with the provision of a set of available features. Apart from the use of the conversational context (contx), we also include the system response (resp), synthetic responses generated by the language model (syn-resp) due to the missing responses as input in the practical scenario, the name entities within the incoming responses (ner), retrieved knowledge (know) and the description of the knowledge source, e.g., the WikiHow website (source). By using various combinations of inputs and customising the corresponding instructions, we explore the effectiveness of the resulting learned language models that implement the RAGate-PEFT.


\textit{\textbf{RAGate-MHA: }} Apart from the use of pre-trained language models and further fine-tuned language models, we also explore the introduction of a multi-head attention neural encoder to model the context as input and estimate the augmentation necessity (i.e., RAGate-MHA). Here, we describe the model structure of RAGate-MHA. At first, as denoted by \cite{vaswani2017attention}, the attention mechanism is formulated as the interaction between three objects, queries $Q$, keys $K$, and values $V$: $Attention(Q,K,V) = \text{softmax}(\frac{QK^T}{\sqrt{d_k}})V$. To estimate the necessity of augmentation, we fit the context and the retrieved knowledge into the roles of these three objects. Specifically, we include the setups of (1) using context only (\textit{contx}) or (2) using the concatenated context and retrieved knowledge (\textit{contx $\oplus$ know}) as queries, keys, and values, and (3) using the context as queries and interact with the retrieved knowledge as keys and values (\textit{contx $\times$ know}).
Next, following \cite{vaswani2017attention} in the encoder construction of a transformer model, we encode the inputs via an input embedding layer into latent vectors and a position encoding layer to encode the order of tokens in the sequence. After that, we leverage the multi-head attention to learn attention weights on the inputs and then followed by a feed-forward network: 
\begin{equation}
    FFN(x) = \text{max}(0, xW_1 + b_1)W_2 + b_2
\end{equation}
where $W_1$ and $W_2$ are two learned parameter matrics with two bias terms ($b_1$ and $b_2$). Both multi-head attention and feed-forward neural modules are followed by residual connection \cite{he2016deep} and layer normalisation \cite{ba2016layer}. Unlike the introduction of another decoder module that addresses the sequence-to-sequence generation in \cite{vaswani2017attention}, we followed the encoder output with a linear projection module and a softmax function for our binary classification task.

\section{Model Training and Evaluation Setups}
We evaluate the performance of introducing RAGate according to its binary classification performance and the effectiveness of the resulting response generation. Specifically, we use the KETOD dataset \cite{chen2022ketod}, which has fully annotated 5,324 dialogues and 52,063 turns of conversations. In particular, it is associated with 33,761 knowledge snippets to be retrieved and augmented. In addition, KETOD was developed with human labels on turns of conversations (around 12.1\% of turns) about the need for augmenting with retrieved knowledge snippets for a natural and informative system response. Hence, we use these human labels as natural ground truths when evaluating RAGate. It is worth indicating that many current knowledge-augmented conversational datasets often ground their conversations on the knowledge snippet, such as Wizard of Wikipedia \cite{dinan2018wizard} and CMU\_DoG \cite{zhou2018dataset}, which makes them not a natural fit to be investigated in this study.

\begin{table}
    \centering
    \resizebox{0.4\textwidth}{!}{
    \begin{tabular}{l|c|c}
    \hline
        \textbf{Retrieval Models} & \textbf{Recall@1} & \textbf{Recall@3} \\
        \hline
        TF-IDF   &  0.0227 & 0.0871 \\
        BERT-Ranker   & 0.2475 & 0.4714 \\
        \hline 
    \end{tabular}}
    \caption{Retrieval Performance Evaluation when using context as the query.}
    \label{tab:retrieval_ranker}
\end{table}

Due to the limited computational resource availability, we explore the use of Llama-v2-7B and Llama-v2-13B to implement RAGate-prompt and fine-tune Llama-v2-7B for RAGate-PEFT. We implement QLoRA using the PEFT library~\cite{peft} and set the lower rank to 16. As discussed in Section~\ref{sec:methodology}, we have various input features to be combined for performance optimisation. We begin with the use of context only, then concatenate the context with the real response (contx-resp), with the synthetic response and recognised entities (contx-syn-resp-ner) and further extend with the use of retrieved knowledge (contx-syn-resp-ner-know) or the source of knowledge (contx-syn-resp-ner-source). Specifically, we retrieve the relevant knowledge by exploring the use of TF-IDF and a learned BERT ranker. We evaluate their performance with the classic Recall@1 and Recall@3 on the test collection. We use a shallow cutoff because we only use top-relevant knowledge snippets for augmentation. Table~\ref{tab:retrieval_ranker} shows their retrieval performance. According to the leading performance of BERT-Ranker, we augment knowledge with its retrieved top 3 relevant knowledge snippets (i.e., $k=3$). Regarding the development of RAGate-MHA, we explore the combinations of 2 to 8 layers, 2 or 4 heads and the embedding size in [64, 128, 256] for the best classification accuracy. We report the precision, recall, F1, Area Under Curve (AUC) and the False Discovery Rate (FDR) as the main measures to show the classification effectiveness. 

Next, we further deploy the best-performing RAGate gate function to update the KETOD dialogue system \cite{chen2022ketod}, which uses GPT-2 \cite{radford2019language} as the backbone model. To highlight the effect of various augmentation setups, we use the context with the gold action without extra prediction as input to KETOD. Then, we compare the resulting performance to the KETOD model without knowledge augmentation and augmenting every system response as baselines. To report the response generation effectiveness, we report how close the response is to the ground truth via BLEU, ROUGE-1/2/L and BERTScores and the confidence score calculated by the minimum probabilities of individual tokens that compose the response.
As argued by \citet{varshney2023stitch}, this calculated confidence score can highly correlate with a language model's likelihood of generating hallucinated responses.  We trained our models and conducted the evaluations on one machine with one NVIDIA 4090 GPU. 

\section{Results and Analysis}

\begin{table}
    \centering
    \resizebox{0.5\textwidth}{!}{
    \begin{tabular}{|lccc|}
    \toprule
    \rowcolor{yellow!25}\textbf{Model Variants}  &  \textbf{Precision} & \textbf{Recall} & \textbf{F1} \\
    \toprule
    \rowcolor{gray!30}
    \multicolumn{4}{c}{\textbf{RAGate-Prompt}: LLMs -- Zero Shot} \\
    \midrule
    Llama-2-7B &  0.1323 & 0.0278 & 0.0460 \\
    Llama-2-13B & 0.1422 & 0.1083 & 0.1230  \\
    \midrule
    \rowcolor{gray!30}\multicolumn{4}{c}{\textbf{RAGate-Prompt}: LLMs -- In-Context Learning} \\
    \midrule
    Llama-2-7B & 0.1417 & 0.0294 & 0.0487 \\
    Llama-2-13B & 0.0989 & 0.0851 & 0.0915 \\
    \midrule
    \rowcolor{gray!30}\multicolumn{4}{c}{\textbf{RAGate-PEFT:}  Parameter Efficient Fine-tuned LLMs (Llama2-7B)} \\
    \midrule
    {[contx$\oplus$resp]} & 0.4926 & 0.3095 & 0.3802 \\
    \midrule
    contx-only & 0.5203 & 0.3359 & \textbf{0.4082} \\
    \midrule
    contx-(syn-resp)-ner &  \textbf{0.6818}  & 0.2321 & 0.3464 \\
    contx-(syn-resp)-ner-know &  0.4698  & 0.0603 & 0.1069 \\
    contx-(syn-resp)-ner-source &  0.4000  & 0.0185 & 0.0355 \\
    \midrule
    \rowcolor{gray!30} \multicolumn{4}{c}{\textbf{RAGate-MHA:} Context with / without Knowledge Input} \\
    \midrule
    MHA(contx)-h(4)-l(5)-emb(64) & 0.3210 & 0.5541 & 0.4065 \\
    MHA([contx$\oplus$know])-h(4)-l(2)-emb(64) & 0.2795 & 0.5201 & 0.3636 \\
    MHA(contx$\times$know)-h(4)-l(2)-emb(64) & 0.2272 & \textbf{0.5835} & 0.3271 \\ 
    \midrule
    \rowcolor{gray!30}\multicolumn{4}{c}{\textbf{RAGate-MHA:} Context-Response Input} \\
    \midrule
    MHA([contx$\oplus$resp])-h(4)-l(4)-emb(64) & 0.3500 & 0.5510 & 0.4281\\
    \bottomrule
    \end{tabular}}
    \caption{Classification accuracy on adaptive augmentation for system response. "contx", "resp",  and "know" refer to the use of context, initial system response, and retrieved knowledge snippets as input. "syn-resp" and "ner" are the additional synthetic response and name entity recognition steps in the model fine-tuning prompts. \textit{h}, \textit{l} and \textit{emb} refer to the best-performed configuration on the number of heads, layers and embedding size.}
    \label{tab:classification}
\end{table}
\subsection{Augmentation Need Classification}
First, we evaluate the classification accuracy of our developed RAGate gate methods for addressing the adaptive RAG to system responses. Table~\ref{tab:classification} presents the classification performance of RAGate baselines while evaluated on the test collection of the KETOD dataset, which includes rich human labels on the use of RAG for response generation. As discussed in Section~\ref{sec:methodology}, we explore the development of RAGate with three variants: the use of LLM prompting (RAGate-Prompt), parameter-efficient fine-tuned LLMs (RAGate-PEFT), and a neural classifier with Multi-Head Attention structure (RAGate-MHA). 

\textbf{\textit{RAGate performance with LLM prompting versus fine-tuning.}} By comparing the corresponding performance reported in Table~\ref{tab:classification}, we observe that, on average, fine-tuning a Llama-2-7B with QLoRA (i.e., RAGate-PEFT) can significantly outperform RAGate-Prompt. For example, by looking at the RAG-PEFT with context-only input, without using extra input features and instruction updates, it can outperform all RAG-Prompt approaches by a big margin (e.g., 0.4082 versus the highest 0.1230 F1 scores). This reflects the difficulty of this adaptive knowledge augmentation task, which can not be properly addressed by prompting a general pre-trained language model. In particular, the use of larger language models and the in-context learning setup, which often result in improved performance~\cite{arora2022ask}, can not guarantee the enhancement of models' classification accuracy regarding this classification task.

Regarding the performance of RAGate-PEFT approaches, by first examining the effect of using synthetic response and recognised name entities, we observe significantly improved precision (0.5203 to 0.6818) but with the cost of lower recall (0.3359 to 0.2321). In addition, when we add the retrieved knowledge to the input features for prediction, we observe a significant performance drop across all evaluated aspects.  This can be caused by the additional complexity introduced by the included retrieved knowledge snippets. Furthermore, we also explored the performance impact of naming the source of the knowledge snippet. We use wikiHow\footnote{https://www.wikihow.com} in this study, which can provide rich task instructions for offering informative task-oriented system response \cite{sen2023task2kb}. However, the fine-tuned model cannot reasonably connect the promised rich resource from the knowledge source and the prediction of augmentation necessity.

\textbf{\textit{RAGate Performance between fine-tuned LLM and MHA classifier.}} Next, by comparing the experimental results of RAGate-MHA and RAGate-PEFT in Table~\ref{tab:classification}, we observe a wide-margin recall improvement using RAGate-MHA, reaching a minimum recall of 0.52, but with significantly lower precision accuracy. In Table~\ref{tab:classification}, we also include the use of both the context and the initial system responses (i.e., MHA([contx, resp])) for additional insights. We can observe that a higher precision can be achieved but the use of response does not improve the recall performance. These results are consistent with the observed performance of RAGate-PEFT, which further encourages the use of a synthetic response due to the unavailability of a system response in a practical scenario. In addition, we also observe a similar performance drop when including the retrieved knowledge snippets for classification. Even though the RAGate-MHA model, using the interaction between context and retrieved knowledge snippets, can achieve the highest recall of 0.5835, it can not outperform the RAGate-MHA using context-only on other metrics. Hence, considering the similar F1 and AUC performance levels of RAGate-PEFT and RAGate-MHA leads to a trade-off balance between precision and recall for the two groups of approaches. To further evaluate the classification effectiveness of RAGate, in Appendix \ref{sec:ragate-userstudy}, we provide a detailed discussion of a conducted user study that explores whether RAGate can also assess the potential contribution of retrieved snippets when predicting the decision for retrieval augmentation.

\begin{figure}
    \centering
    \includegraphics[width=.95\columnwidth]{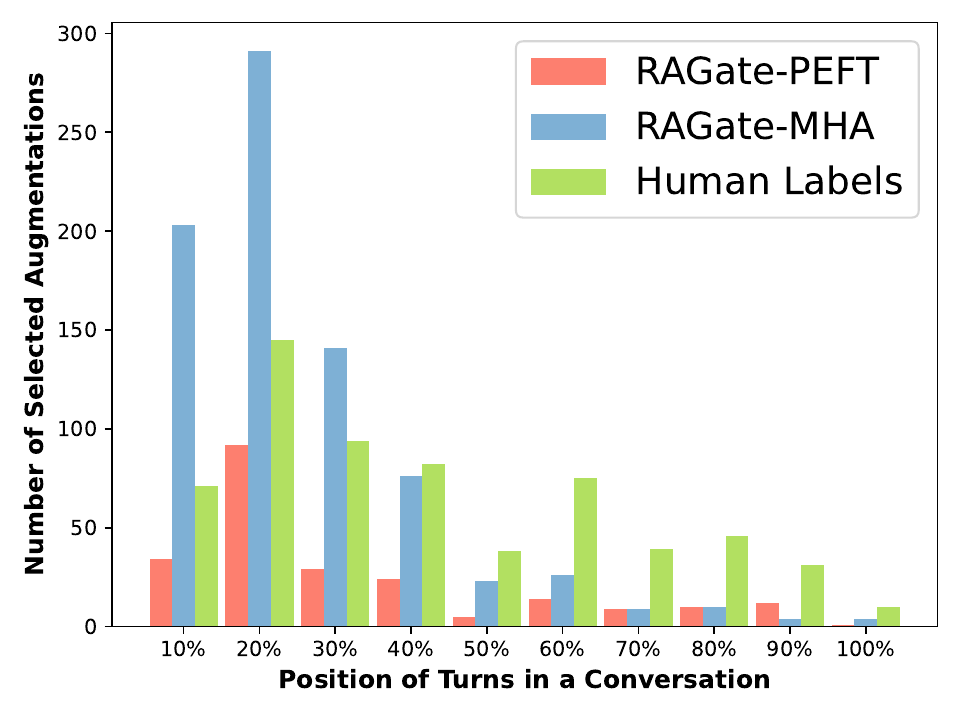}
    \caption{Frequency analysis of adaptive augmentations about the position of a conversation.}
    \label{fig:position_freq}
\end{figure}

\subsection{Adaptive Augmentation Analysis}
In addition to the classification accuracy, we also compare the choice of human workers and RAGate approaches in augmenting specific turns. Specifically, we analyse the frequency of augmentation in different positions of conversations and different domains covered in the KETOD dataset. We use the RAGate-PEFT (contx-(syn-resp)-ner) with the highest precision and RAGate-MHA (MHA(contx)) with the best overall performance in the above analysis as representatives for comparison. Figure~\ref{fig:position_freq} presents the frequency in different positions. Due to the unequal number of conversational turns, we use the ratio to indicate the relative position. According to the reported results in Figure~\ref{fig:position_freq}, most human augmentation selections happen at the beginning of a conversation. This trend is also effectively captured by both RAGate approaches, especially RAGate-MHA. This can be caused by the reason that a conversation is semantically coherent, and once sufficient additional information is provided at the early stage, the value of knowledge augmentation to the later turns is naturally lower. 

\begin{figure}
    \centering
    \includegraphics[width=\columnwidth]{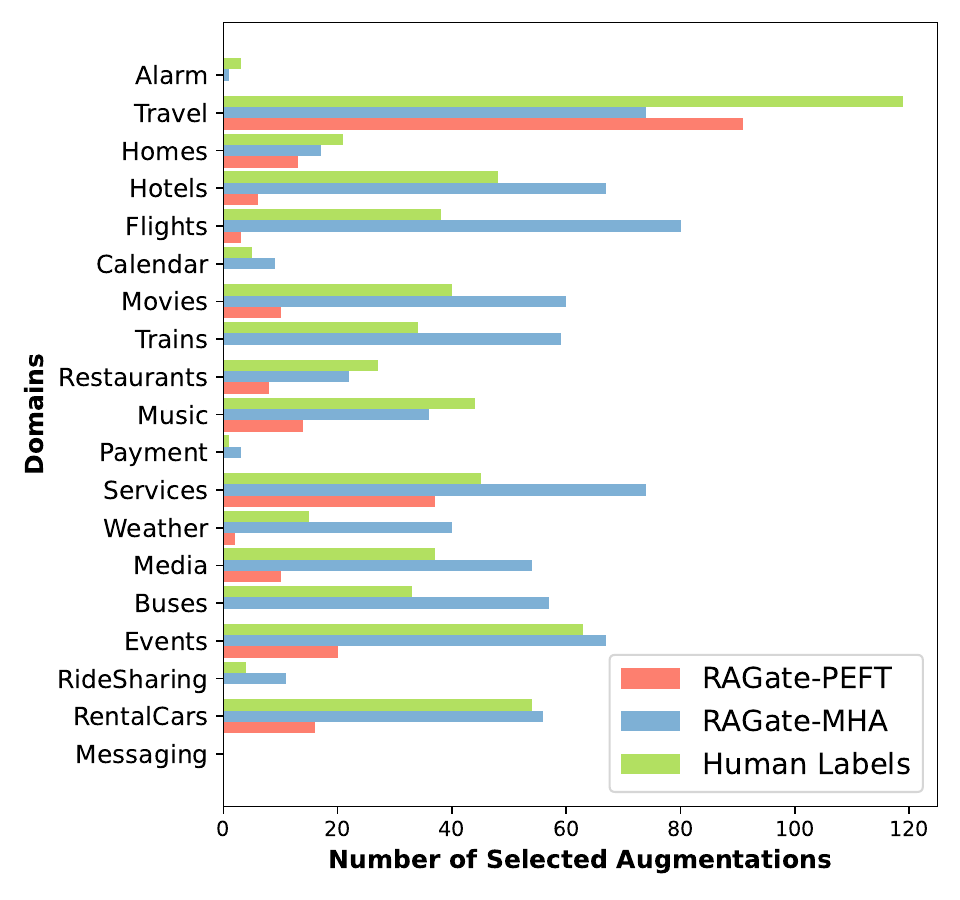}
    \caption{Frequency analysis of adaptive augmentations about dialogue domains.}
    \label{fig:domain_freq}
\end{figure}
On the other hand, Figure~\ref{fig:domain_freq} presents the augmentation frequency over different domains. We observe that system responses about certain domains are selected more often by humans than other domains, such as travel, hotels, trains, flights, service and rental cars, which require access to additional information to assist the \textit{suggestion-making}, and the domains, like movies, music, media, events that often include entities require \textit{enriched description}. By looking into the performance of RAGate-PEFT and RAGate-MHA, RAGate-MHA can make aligned selections for humans. However, the RAGate-PEFT does not guarantee the identification of appropriate augmentation use and often presents fewer augmentations, apart from the travel domain. Hence, by considering both position and domain augmentation frequency, we conclude that RAGate-MHA can outperform RAGate-MHA and effectively capture the trend of augmentation needs.  

\begin{table}
    \centering
    \resizebox{.5\textwidth}{!}{%
    \begin{tabular}{|lcllllc|}
    \toprule
     \rowcolor{yellow!25}\textbf{Variants }& \textbf{\# Augs} & \textbf{BLEU} & \textbf{ROUGE-L} & \textbf{BERTScore} & \textbf{Confidence} & \\
     \toprule
     \rowcolor{green!10}\textbf{No-Aug} & 0 & 9.38$\bullet$ & 0.3780$\bullet$ &  0.8105$\bullet$ & 9.3425$\bullet$ & -- \\
     \midrule
     \rowcolor{gray!30}\multicolumn{7}{|c|}{\textbf{Augment BERT Ranker Retrieved Knowledge}} \\
     \textbf{RAGate-PEFT} & 230 & 10.45  & 0.3825 & 0.8144 & 9.3374 & -0.05\% \\
     \textbf{RAGate-MHA} & 787 & 12.14 & 0.3882 & 0.8192 & 9.3083  & -0.36\%\\
     \rowcolor{gray!10}\textbf{Random-Aug} & 230 & 9.53  & 0.3784& 0.8110& 9.2984 & -0.47\%\\
     \rowcolor{gray!10}\textbf{Random-Aug} & 787 & 10.01  & 0.3795 & 0.8126 & 9.1877 & -1.65\%\\
     \textbf{Human-label} & 631 & 11.66 & 0.3856 & 0.8176 & 9.2550 & -0.93\%\\
     \rowcolor{violet!10}\textbf{Aug-All} & 4964 & 16.08 & 0.3927& 0.8258 & 8.3677 & -10.43\%\\
     \midrule
     	

    \rowcolor{gray!30}\multicolumn{7}{|c|}{\textbf{Augment Rank-1 Relevant Knowledge}} \\
    \midrule
    \textbf{RAGate-Llama} & 230 & 10.54  & 0.3822 & 0.8142 & 9.3642 & +0.23\%\\
    \textbf{RAGate-MHA} & 787 & 11.99  & 0.3883 & 0.8191 & \textbf{9.3774} & +0.37\%\\
    \rowcolor{gray!10}\textbf{Random-Aug} & 230 & 9.51  & 0.3784& 0.8110 & 9.3328 & -0.10\% \\
    \rowcolor{gray!10}\textbf{Random-Aug} & 787 & 10.01  & 0.3800 & 0.8127 & 9.2982 & -0.47\%\\
     \textbf{Human-label} & 631 & 11.52& 0.3846 & 0.8170& 9.3218 & -0.22\% \\
    \rowcolor{violet!10}\textbf{Augment-All} & 4964 & 16.05 & 0.3944 & 0.8259 & 9.0655 & -2.9\%\\
    \bottomrule
    \end{tabular}%
    }
    \caption{Performance of applying RAGate and compared to the KETOD baseline on the KETOD dataset. Confidence is calculated by the average value over the lowest logit of each generation.}
    \label{tab:generation}
\end{table}

\subsection{RAGate for Response Generation}
To evaluate the effect of adaptive RAG for a conversational system, we use RAGate-PEFT (contx-(syn-resp)-ner) with the highest precision and RAGate-MHA (MHA(contx)) with the best overall performance in the above analysis, to support the adaptive retrieval augmented conversational response generation. Table \ref{tab:generation} presents the results of applying RAGAte to the KETOD model for adaptive knowledge augmentation when evaluated on the KETOD dataset. We include four types of adaptive augmentation, namely the use of RAGate and comparison to the random selection with equal numbers of selections, human choice, and the commonly used "all" augmentation. 
In addition, to explore the effect of varied quality of knowledge snippets, we also extend the evaluation of using the top-3 knowledge snippets ranked by different retrievers (i.e., BERT-ranker and TF-IDF) and the use of knowledge snippets at the 1st and 5th rank according to the BERT-ranker. Due to the space limit, we first present the results of using BERT-ranker retrieved and top-1 relevant knowledge and top-1 relevant in Table~\ref{tab:generation} and show the full results in the Appendix~\ref{sec:additional_results}. 

At first, without adaptive knowledge augmentation, we compare the choice of response generation without augmentation and with "always" augmentation (i.e., No-Aug versus Aug-All). In Table \ref{tab:generation}, we observe that by augmenting a total of 4,964 system responses in the test collection, the conversational model can generate more informative and effective responses according to the reported scores of BLEU, ROUGE and BERTscore. This aligns with the reported effectiveness of RAG in many existing studies. However, we also identify a significant drop in the model's generation confidence level. As denoted by \citet{varshney2023stitch}, a lower confidence level can correlate with a higher chance of generating hallucinated responses, which could be caused by the unnecessary use of external knowledge. Hence, to investigate the effectiveness of adaptive knowledge augmentation, we examine the impact of using RAGate. According to the reported experimental results in Table \ref{tab:generation}, the adaptive augmented response generation with fewer knowledge snippets can indeed result in a higher confidence level than Aug-All. 

Moreover, comparing the performance between RAGate and random selections shows that, considering equal numbers (230 or 787 according to the classification with RAGate) of system responses for augmentation, RAGate can further result in a higher quality of generated response. RAGate-MHA even enables results that are comparable to Aug-All's response quality, with only 787 turn augmentations instead of all 4964 turns. Specifically, the use of RAGate-PEFT, which identifies 230 turns of system responses for knowledge augmentation, can even outperform the random baseline that augments 787 system response turns with improved response quality. 
Apart from the improved response quality, RAGate also enables the conversational model to maintain a high confidence level and ensure faithful responses. Indeed, using RAGate-MHA, which augments 787 system responses, only lowers the average confidence score by 0.36\%, instead of the 1.65\% when randomly selecting an equal number of turns to augment. 

In addition, considering the use of different quality and amount of knowledge snippets for augmentation, we also include the use of the most relevant knowledge snippet according to BERT-ranker in Table~\ref{tab:generation}. We observe that the use of different amounts of knowledge snippets in different relevance levels has a marginal effect on this learned dialogue system. However, we observe a significant difference in the confidence level. We observe that using only the most relevant knowledge snippet enables the Aug-All to suffer less from a lower confidence level. In particular, the application of RAGate can even increase the confidence level of the conversation system in response generation. This indicates that the confidence score can also correlate with the quality of the augmented knowledge snippets. This observation is further validated using knowledge snippets with fifth-ranking positions by BERT-ranker and the use of TF-IDF ranker.  
We include the full experimental results in Table~\ref{tab:generation-appendix} and attached in the Appendix. 
These observations indicate the value of adaptive system response augmentation via RAGate in generating high-quality outputs, ensuring faithful responses, and potentially saving retrieval costs. We also show the value of using confidence scores to reflect the contribution of RAG.

\section{Conclusions}
Our study investigates a core research question about whether retrieval-augmented generation is always useful to a conversational system. To answer this research question, we propose adaptive retrieval-augmented generation for conversational systems and introduce corresponding gate functions, RAGate, for explicit control. A comprehensive set of experiments and results show the RAGate approaches can effectively identify augmentation needs. In addition, RAGate can capture human preference by augmenting the beginning turns of conversations, and RAGate can further identify knowledge augmentation for assisting suggestion-making and enriching description. When applying RAGate to conversational systems, we observe that it can ensure comparable quality of generated responses and enable the system to increase generation confidence for faithful outputs, especially with the appropriate use of relevant knowledge snippets.

\newpage

\section*{Limitations}
There are three limitations of this study. At first, due to the main focus of examining the adaptive retrieval-augmented generation for a conversation system. We only consider a few examples of retrieval techniques (TF-IDF and BERT-ranker), which can be further extended to recent retrieval techniques, such as dense passage retrieval for additional insights. The second limitation is the missing use of larger language models, such as GPT-4, due to the shortage of computational resources. Including larger language models for conversational systems could introduce additional experimental insights. The third limitation is the shortage of appropriate conversational data for extensive evaluations. This is mainly caused by the recent development of the retrieval augmented generation technique and its application to conversational systems. Future research is encouraged to address this limitation. 




\section*{Ethics Statement}
All experiments in this study were conducted using publicly available datasets and open-released language models, which do not contain any private information that could raise ethical concerns. 

\bibliography{custom}

\appendix

\section{Prompts for RAGate-Prompt}
\label{sec:ragate-prompt}
In this section, we list the used prompts for the RAGate-Prompt gate mechanism. 

\textbf{Zero-Shot Prompt: }

Below is an instruction that describes a task. Please respond with `True' or `False' only that appropriately completes the request. 

\#\#\# Instruction: Analyse the conversational context so far. Generate an appropriate response. Consider the invovled entites. Estimate if augmenting the response with external knowledge is helpful with an output of `True' or `False' only.

\#\#\# Input: [Converstion Context Input]

\#\#\# Response: 

\begin{table*}
    \centering
    \resizebox{.8\textwidth}{!}{%
    \begin{tabular}{|lccccccc|}
    \toprule
     \rowcolor{yellow!25}\textbf{Augmentation Variants }& \textbf{\# Augs} & \textbf{BLEU} & \textbf{ROUGE-1} & \textbf{ROUGE-2} & \textbf{ROUGE-L} & \textbf{BERTScore} & \textbf{Confidence} \\
     \toprule
     \rowcolor{green!10}\textbf{No-Aug} & 0 & 9.38 & 0.4111 & 0.2246 & 0.3780 &  0.8105 & 9.3425 \\
     \midrule
     \rowcolor{gray!30}\multicolumn{8}{|c|}{\textbf{Augment BERT Ranker Retrieved Knowledge}} \\
     \textbf{RAGate-Llama} & 230 & 10.45 & 0.4165 & 0.2273 & 0.3825 & 0.8144 & 9.3374\\
     \textbf{RAGate-MHA} & 787 & 12.14 & 0.4227 & 0.2318 & 0.3882 & 0.8192 & 9.3083 \\
     \rowcolor{gray!10}\textbf{Random-Aug} & 230 & 9.53 & 0.4119 & 0.2250 & 0.3784 & 0.8110 & 9.2984 \\
     \rowcolor{gray!10}\textbf{Random-Aug} & 787 & 10.01 & 0.4138 & 0.2265 & 0.3795 & 0.8126 & 9.1877 \\
     \textbf{Human-label} & 631 & 11.66	& 0.4198 & 0.2297 & 0.3856 & 0.8176 & 9.2550 \\
     \rowcolor{violet!10}\textbf{Augment-All} & 4964 & 16.08	& 0.4301 & 0.2364 & 0.3927 & 0.8258 & 8.3677 \\
     \midrule
     	
    \rowcolor{gray!30}\multicolumn{8}{|c|}{\textbf{Augment TF-IDF Ranker Retrieved Knowledge}} \\
     \textbf{RAGate-Llama} & 230 & 10.52 & 0.4165 & 0.2273 & 0.3826 & 0.8144 & 9.3418\\
     \textbf{RAGate-MHA} & 787 & 12.11 & 0.4233 & 0.2319 & 0.3889 & 0.8193 & 9.3058 \\
    \rowcolor{gray!10}\textbf{Random-Aug} & 230 & 9.47 & 0.4118 & 0.2251 & 0.3783 & 0.8110 & 9.3006 \\
    \rowcolor{gray!10}\textbf{Random-Aug} & 787 & 9.93 & 0.4136 & 0.2259 & 0.3793 & 0.8125 & 9.1942 \\
    \textbf{Human-label} & 631 & 11.60	& 0.4198 & 0.2293 & 0.3854 & 0.8175 & 9.2639 \\
    \rowcolor{violet!10}\textbf{Augment-All} & 4964 & 15.76	& 0.4289 & 0.2345 & 0.3914 & 0.8256 & 8.4188 \\
    \midrule

    \rowcolor{gray!30}\multicolumn{8}{|c|}{\textbf{Augment Rank-1 Relevant Knowledge}} \\
    \midrule
    \textbf{RAGate-Llama} & 230 & 10.54 & 0.4162 & 0.2271 & 0.3822 & 0.8142 & 9.3642\\
    \textbf{RAGate-MHA} & 787 & 11.99 & 0.4227 & 0.2316 & 0.3883 & 0.8191 & \textbf{9.3774} \\
    \rowcolor{gray!10}\textbf{Random-Aug} & 230 & 9.51 & 0.4117 & 0.2250 & 0.3784 & 0.8110 & 9.3328 \\
    \rowcolor{gray!10}\textbf{Random-Aug} & 787 & 10.01 & 0.4140 & 0.2267 & 0.3800 & 0.8127 & 9.2982 \\
     \textbf{Human-label} & 631 & 11.52	& 0.4189 & 0.2289 & 0.3846 & 0.8170 & 9.3218 \\
    \rowcolor{violet!10}\textbf{Augment-All} & 4964 & 16.05	& 0.4308 & 0.2365 & 0.3944 & 0.8259 & 9.0655 \\
    \midrule
    \rowcolor{gray!30}\multicolumn{8}{|c|}{\textbf{Augment Rank-5 Relevant Knowledge}} \\
    \midrule
    \textbf{RAGate-Llama} & 230 & 10.47 & 0.4161 & 0.2272 & 0.3823 & 0.8142 & 9.3592\\
    \textbf{RAGate-MHA} & 787 & 12.18 & 0.4224 & 0.2314 & 0.3883 & 0.8192 & 9.3704 \\
    \rowcolor{gray!10}\textbf{Random-Aug} & 230 & 9.52 & 0.4118 & 0.2252 & 0.3785 & 0.8110 & 9.3315 \\
    \rowcolor{gray!10}\textbf{Random-Aug} & 787 & 10.01 & 0.4135 & 0.2263 & 0.3794 & 0.8127 & 9.2961 \\
    \textbf{Human-label} & 631 & 11.58	& 0.4186 & 0.2287 & 0.3845 & 0.8170 & 9.3210 \\
    \rowcolor{violet!10}\textbf{Augment-All} & 4964 & 15.97 & 0.4290  & 0.2349 & 0.3927 & 0.8256 & 9.0604\\
    \bottomrule
    \end{tabular}%
    }
    \caption{Performance of applying RAGate and compared to KETOD on the SGD dataset. Confidence is calculated by the average value over the lowest logit of each generation.}
    \label{tab:generation-appendix}
\end{table*}

\textbf{In-Context Learning Prompt: }

Below is an instruction that describes a task. Please respond with `True' or `False' only that appropriately completes the request. 

\#\#\# Instruction: Analyse the conversational context so far. Generate an appropriate response. Consider the invovled entites. Estimate if augmenting the response with external knowledge is helpful with an output of `True' or `False' only.

\#\#\# Example 1: USER: I'm planning a trip, can you help me look for a flight? SYSTEM: Which day are you planning to return and from which city? USER: I want to go from NYC the day after tomorrow and return on the 13th of this month. SYSTEM: Where would you like to go? USER: I want to go to Vancouver, BC. Can you look for a Premium Economy class ticket. SYSTEM: I found 1 flight for you. It is a Delta Airlines flight that takes off at 6 am and returns at 2:50 am. The price is \$505. USER: What is the departure airport, and how many stops does the flight have?

\#\#\# Response: True

\#\#\# Example 2: USER: Get me bus tickets to a Cher event on March 6th. SYSTEM: How many to buy? USER: only one, please.

\#\#\# Response: False

\#\#\# Input: [Converstion Context Input]

\#\#\# Response: 

\section{Impact of Retrieval Quality on Adaptive RAG}\label{sec:ragate-userstudy}
To have a successful conversation model with a retrieval-augmented system, two main criteria must be met. One is identifying insufficient context, and the other is the quality of retrieved information \cite{salemi2024evaluating, yu2024evaluation}. A conversational model performs better when both criteria are satisfied. In our proposed approach, as shown in Table 2, we have already assessed whether our adaptive retrieval method can detect insufficient context. We further explored to determine whether our model can inherently estimate the quality of the retrieved snippets to address such insufficiency and, based on that, decide on the retrieval. Although we do not explicitly provide retrieved snippets to our model, retrieval comes with a corpus that includes potentially relevant knowledge snippets. Consequently, given a query and a retrieval collection, it can be estimated whether useful information for the query exists in the corpus to address the insufficient context. To investigate by following this direction, we randomly selected $50$ samples from instances where our proposed approach (RAGate-MHA, the best-performing gate model) predicted using retrieval augmentation. We asked domain experts (co-authors) to score whether they thought the retrieved snippets in those scenarios could be useful to response generation. Users rated the snippets on a scale of $0-4$, with scores of $3$ or $4$ indicating `useful' or `highly useful'. We found that in $54\%$ of cases where the prediction was for augmentation, users also found the snippets useful. This indicates that our proposed approach can implicitly capture the potential for obtaining high-quality retrieval snippets. 

\section{Additional experimental results about RAGate for Response Generation} \label{sec:additional_results}
In Table~\ref{tab:generation-appendix}, we include the complete experimental results of applying RAGate for adaptive retrieval-augmented system response generation. Specifically, explore the use of retrieved knowledge snippets to different extents of relevance. We include top-3 knowledge snippets retrieved by BERT-ranker and TF-IDF. In addition, we also explore the use of knowledge snippets in different ranking positions (rank 1 and 5) according to the BERT-ranker retriever. The experimental result shows that precisely using a suitable amount of relevant knowledge can generate a response with higher confidence (i.e., less is more). In addition, this observation also indicates the potential use of confidence levels to evaluate the quality of the augmented knowledge.

\end{document}